\documentclass[letterpaper]{article} 
\usepackage{aaai2027}  
\usepackage[hyphens]{url}  
\usepackage{graphicx} 
\usepackage{natbib}  
\usepackage{caption} 
\usepackage{algorithm}
\usepackage{algorithmic}
\usepackage{amsmath, amsthm}
\usepackage{subcaption}
\usepackage{soul}

\newtheorem*{problem*}{Example Problem}
\usepackage{newfloat}
\usepackage{listings}
\DeclareCaptionStyle{ruled}{labelfont=normalfont,labelsep=colon,strut=off} 
\floatstyle{ruled}
\newfloat{listing}{tb}{lst}{}
\floatname{listing}{Listing}

\usepackage{booktabs}

\title{RamseyGadgets: A Graph Construction Dataset for LLMs\thanks{Supported in part by NSF grant CCF-2421977 and DUE-2439323.}}
\author{
    Zohair Raza Hassan\equalcontrib\corresponding, Deepak Pandita\equalcontrib\corresponding
}
\affiliations{
    Rochester Institute of Technology, NY, USA \\
    zh5337@rit.edu, deepak@mail.rit.edu
}

\nocopyright

\begin{document}

\maketitle

\begin{abstract}
    Constructing special graphs is an important task within graph theory and computer science. Many popular graph constructions are the result of a comprehensive exploration of relevant graphs and human ingenuity. Given the rise of generative AI usage in mathematics, it is natural to test whether LLMs are able to construct graphs with specified properties using their reasoning capabilities.
Unfortunately, many natural graph construction problems, such as finding extremal Ramsey-good graphs (i.e., avoiding specific monochromatic subgraphs), have been explored extensively in the literature, making it difficult to ascertain whether a construction is the product of an LLM's reasoning capabilities or its recollection from training data. 
In this work, we introduce \textbf{RamseyGadgets}, a novel dataset of 70 underexplored graph construction problems that require finding Ramsey-good graphs with special properties (e.g., containing an edge with a fixed color). These problems have reasonably sized solutions (at most 10 vertices) that can be verified by SAT solvers, making them suitable for automatic evaluation. Our dataset is easily expandable, as one can simply change the monochromatic subgraphs being avoided to obtain a new set of problems. 
We evaluate the performance of five open-source LLMs on our dataset and report the results. Our findings show that LLMs achieve only 37.70\% accuracy on the hard-tier problems in our dataset, with Gemma-4-31B achieving the highest performance out of the five.
We also showcase how our dataset allows us to ascertain what kind of hints help LLMs perform better at this task. 
\end{abstract}


\section{Introduction}

Constructing graphs with special properties is a task frequently faced by researchers in mathematics and computer science. 
Popular examples include constructing 
extremal graphs in Ramsey theory~\cite{ds1}, expander graphs for derandomization~\cite{DBLP:books/daglib/0023084}, and graph gadgets for NP-hardness proofs~\cite{DBLP:books/fm/GareyJ79}.
Constructing special graphs is an especially laborious and creative exercise;
such graphs are typically found
after the careful inspection of several graphs, through which 
researchers eventually discover a pattern that they can exploit to obtain a graph with the desired properties. 
Constructions that are the product of human ingenuity can have lasting effects that generalize to other problems.
For example, despite their age, the celebrated graphs of Petersen~\citeyearpar{petersen1898sur} and Schläfli~\citeyearpar{schlafli1858attempt} and their generalizations are still used to solve problems today~\cite{DBLP:conf/soda/InoueKMMS26,DBLP:journals/dam/GoedgebeurO22}.
As we move towards our new age of AI-assisted mathematics,
it is natural to want to use Large Language Models (LLMs) for these arduous yet frequently encountered and fruitful tasks.

\begin{figure*}
    \centering
    \includegraphics[width=0.9\linewidth]{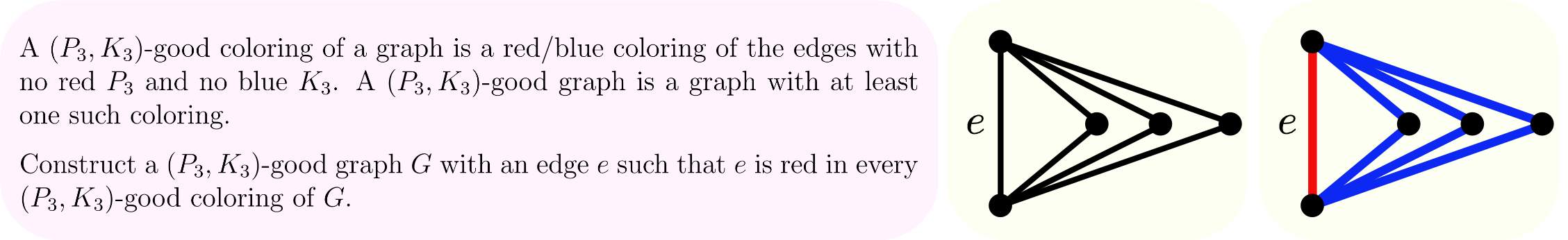}
    \caption{\textbf{Left:} an example of a problem in \textbf{RamseyGadgets}. \textbf{Right:} a solution to the problem and a coloring adhering to the constraints described in the problem.}
    \label{fig:intro-example}
\end{figure*}

Evaluating and training LLMs for this endeavor requires a high-quality dataset. While datasets for general mathematical and algorithmic problems exist (see Section~\ref{sec:rw} for an overview), one specific to graph construction does not. 
Datasets specific to a type of math problem would not only provide more insight into an LLM's limitations for said type of problem, but also allow us to ask new questions. We expand on this concept later. 
Several challenges arise when attempting to build such a specialized dataset. 
To avoid recollection from training data, we need problems that have received little attention in the literature. Many natural graph construction problems, such as those within Ramsey theory, have been explored extensively, and most open cases require large constructions~\cite{ds1} that would be difficult to work with within a reasonable context length or verify efficiently. Another important factor is scalability: as the performance of LLMs improves, datasets become saturated~\cite{DBLP:conf/icml/BalunovicD0PV25} and new problems are needed to evaluate LLM performance.

In this work, we introduce \textbf{RamseyGadgets}, a novel dataset of 70 graph construction problems that addresses all of these challenges.
The dataset is based on finding graphs with special Ramsey-good colorings (i.e., edge colorings avoiding forbidden monochromatic subgraphs---see Preliminaries for a formal definition). 
These problems are:
\begin{itemize}
    \item \textbf{Less examined.} Our problems are based on graphs with special Ramsey-good colorings for cases which only have a handful of relevant papers.
    \item \textbf{Reasonably sized.} All of the problems are witnessed by solution graphs with at most 10 vertices.
    \item \textbf{Easily verifiable.} Checking whether a given graph satisfies the properties specified by the problem is verifiable via a simple script using a SAT solver.
    \item \textbf{Scalable.} Ramsey-good colorings are based on avoiding specified forbidden subgraphs. The dataset can be easily expanded by varying these forbidden subgraphs.
\end{itemize}

An example is provided in Figure~\ref{fig:intro-example}. We provide results on the performance of five open-source LLMs on \textbf{RamseyGadgets}, where we also showcase how our specialized dataset allows us to ask and analyze domain-specific research questions, such as:

\begin{itemize}
    \item \textbf{RQ1.} \textit{Does providing a domain-specific hint help LLMs perform better at the task?}
    \item \textbf{RQ2.} \textit{Does performance decrease as minimal solution size grows?}
    \item \textbf{RQ3.} \textit{Are LLMs able to perform better when given access to tools specialized for the task?}
\end{itemize}

Even with access to a special tool, the best accuracy achieved by an LLM is \textbf{51.83\%} on our dataset, with \textbf{37.70\%} accuracy on the hard-tier problems.
Our dataset is publicly available at
\url{https://github.com/deepakpandita57/RamseyGadgets} .
Our contributions are summarized as follows:
\begin{itemize}
    \item We introduce \textbf{RamseyGadgets}, a novel dataset to evaluate LLM reasoning on graph construction problems.
    \item We evaluate the performance of five popular open-source LLMs on \textbf{RamseyGadgets}, showcasing its difficulty.
    \item We demonstrate how our specialized dataset allows us to ask domain-specific research questions about LLMs' reasoning ability.
\end{itemize}

\section{Related Work}
\label{sec:rw}

There has been an increased interest in studying the abilities of LLMs for graph-based applications~\cite{jin2024large}.
\citet{wang2025graph} explore how LLMs can enhance graph machine learning methods and how graphs can be adopted to improve the performance of LLMs.
With the improving reasoning abilities of LLMs, recent work has utilized computational complexity for more accurate and reliable assessment~\cite{fan-etal-2024-nphardeval}.
\citet{fan-etal-2024-nphardeval} introduce NPHardEval, a dynamic reasoning benchmark built around the framework of computational complexity. The benchmark comprises algorithmic questions scaled across different computational complexity classes, including NP-Hard problems, and the problems are procedurally refreshed monthly to ensure models are constantly evaluated on unseen data instances.
\citet{duchnowski-etal-2025-knapsack} introduced a dataset of Everyday Hard Optimization Problems (EHOP) that translate standard NP-hard algorithmic problems (such as graph coloring or the knapsack problem) into natural language to identify gaps in LLM performance and found large performance disparities relative to the well-documented textbook form.
\citet{hazra2025have} utilized the 3-SAT — the prototypical NP-complete problem — to evaluate LLMs, specifically leveraging the 3-SAT ``phase transition'' phenomenon to dynamically control problem hardness.

\citet{heyman2025evaluating} introduced a dataset of graph $k$-coloring problems across varying complexities: 4 to 8 vertices and 2 to 4 colors (e.g., 4v2c up to 8v4c) to investigate the systematic reasoning capabilities of LLMs.
MathConstruct~\cite{DBLP:conf/icml/BalunovicD0PV25} is a benchmark of 127 challenging problems sourced from various mathematics competitions, which targets constructive proofs. Instead of finding a single numerical answer, the task requires the LLM to construct a specific mathematical object such as a set, matrix, or graph that satisfies a given set of properties. These generated objects can be automatically verified using custom evaluators. They found that the models were able to achieve an accuracy of only 53\% on their benchmark.
BeyondBench~\cite{srivastava2026beyondbench} uses algorithmic problem generation to create mathematically grounded problems on the fly, ensuring contamination resistance. Contamination resistance is guaranteed through the vast problem space, strict mathematical verification, and isomorphic transformations that create syntactically new but semantically identical problems. The hard suite in BeyondBench also contains 10 variations of graph coloring problems.

\textbf{RamseyGadgets} is close to MathConstruct in that our work also requires the construction of mathematical objects, but we are specifically focused on graph construction. Our dataset is based on finding graphs with special Ramsey-good colorings (i.e., edge colorings avoiding forbidden monochromatic subgraphs). These graphs are underexplored, reasonably sized, easily verifiable, and scalable.

We note that constructing special graphs computationally is a popular approach~\cite{ds1,wg2026}, but this often requires specialized algorithms based on intricate knowledge of the underlying problem to effectively prune the search-space of the desired graph. 
While LLMs have successfully been used to prune the search-space for genetic algorithms~\cite{DBLP:journals/corr/abs-2603-09172,DBLP:journals/corr/abs-2506-13131,DBLP:journals/corr/abs-2605-01120,openevolve}, this tells us little about their reasoning abilities on problems of this type.
\textbf{RamseyGadgets} bridges this gap by allowing us to ask and analyze domain-specific research questions for graph construction problems.

\begin{figure*}
    \centering
    \includegraphics[width=0.8\linewidth]{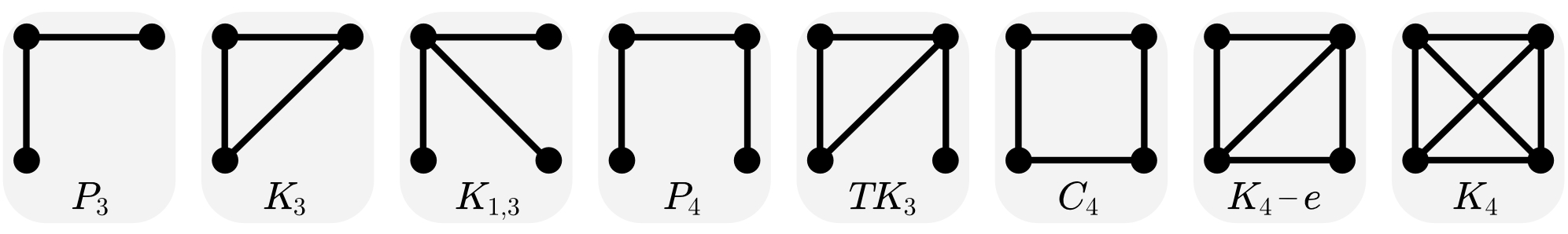}
    \caption{All graphs used as $F$ or $H$ in our dataset.}
    \label{fig:allg}
\end{figure*}

\section{Preliminaries}
\label{sec:prelim}

The notion of Ramsey-goodness concerns coloring the edges of a graph while avoiding specific (not necessarily induced) monochromatic subgraphs. 
Formally, a $(F, H)$-good coloring of a graph is a red/blue edge-coloring containing no red $F$ and no blue $H$, and a graph is called $(F, H)$-good if it has at least one such coloring. 
For example, the coloring of the graph shown in Figure~\ref{fig:intro-example} is a $(P_3, K_3)$-good coloring. The path, cycle, and complete graphs on $n$ vertices are denoted as $P_n$, $C_n$, and $K_n$. The star graph on $n+1$ vertices is denoted as $K_{1,n}$. Note that $P_3 = K_{1,2}$.

The graphs used for $F$ and $H$ in our work are illustrated in Figure~\ref{fig:allg}.
As is standard, we assume that $F$ and $H$ are connected and have at least three vertices (if $F = P_2$ (resp., $H = P_2$), any $(F,H)$-good coloring cannot contain any red (resp., blue) edge). 
Note that a graph is $(F,H)$-good if and only if it is $(H, F)$-good. Since colors can be interchanged without loss of generality, problems on $(F, H)$-good graphs and problems on $(H, F)$-good graphs are equivalent.

\section{RamseyGadgets}

Consider the following graph construction problem:

\begin{problem*}
Construct a $\boxed{(P_3, K_3)\text{-good}}$ graph $G$ with an \ul{edge $e$ such that $e$ is red in every $(P_3, K_3)$-good coloring of $G$.}
\end{problem*}

The problem is asking for a graph adhering to two constraints: (1) the Ramsey-goodness constraint enclosed in $\boxed{\cdots}$, and (2) the underlined ``gadget'' constraint. 
\textbf{RamseyGadgets} consists of 70 similarly formatted graph construction problems that were obtained by varying these constraints.
In our dataset, we use 13 distinct pairs $(F, H)$ and three types of gadgets. We discuss the details of these below.

\subsection{Gadget Types}

The typical goal within Ramsey theory is to find the smallest complete graphs for which no $(F, H)$-good coloring exists.
Less popular, but still important, is 
the search for special graphs (hereafter referred to as gadgets) with restricted colorings known as ``determiner gadgets,'' ``sender gadgets,'' and ''hardness gadgets,'' which are used for constructing families of minimal extremal graphs~\cite{burr1976graphs} and for proving NP-hardness~\cite{Bu3,Scha,wg2026}. 
Since these gadget types are less examined in the literature, they make for good candidates for our dataset. 

\paragraph{Determiner Gadgets.} 
These gadgets are based on forcing an edge to always be a specific color~\cite{Scha}.
For a pair $(F, H)$, a \textbf{$(F,H)$-red-determiner} is a graph $G$ with an edge $e$ such that $e$ is red in every $(F,H)$-good coloring of $G$. \textbf{$(F,H)$-blue-determiners} are defined similarly. 

\paragraph{Sender Gadgets.} 
These gadgets are based on ``sending signals'' across two edges (see~\cite{burr1976graphs}). For a pair $(F, H)$, a \textbf{$(F,H)$-positive-sender} is a graph $G$ with distinct edges $e$ and $f$ such that $e$ and $f$ are the same color
in every $(F,H)$-good coloring of $G$. Moreover, there must exist a good coloring where $e$ is red and a good coloring where $e$ is blue\footnote{Note that without this restriction we could simply take the disjoint union of two determiner gadgets to construct a ``sender.''}.
\textbf{$(F,H)$-negative-senders} are defined similarly, but $e$ and $f$ must always be opposite colorings.

\paragraph{Hardness Gadgets.} 
These gadgets were introduced to simulate clauses and variables in SAT formulas~\cite{Scha,hassan2024,wg2026}. In~\cite{wg2026}, these are defined explicitly for $F = P_3$, and it is noted that the definitions can be extended to any $F$ that is a tree.
We provide the definitions for $F = K_{1,k}$, since that is the only case for $F$ considered in our work (see the preceding section for more details).
\begin{itemize}
    \item In a $(K_{1,k}, H)$-good coloring, an enforced vertex is a vertex incident to no red edges.
    A \textbf{$(K_{1,k}, H)$-clause-gadget} is a $(K_{1,k}, H)$-good graph with three vertices $i_1$, $i_2$, and $i_3$, such that: (1) $G$ does not have a $(K_{1,k}, H)$-good coloring where $i_1$, $i_2$, and $i_3$ are all simultaneously enforced, and (2) $G$ does have $(K_{1,k}, H)$-good colorings for all other 7 combinations of enforcement for $i_1$, $i_2$, and $i_3$.
    \item In a $(K_{1,k}, H)$-good coloring, an enforcer vertex is a vertex incident to $k-1$ red edges. A \textbf{$(K_{1,k}, H)$-variable-gadget} is a graph with vertices $u_1$, $u_2$, and $n_1$ such that:
    \begin{enumerate}
        \item In every $(K_{1,k}, H)$-good coloring of $G$, if $u_1$ or $u_2$ is not an enforcer vertex, then $n_1$ must be an enforcer vertex.
        \item In every $(K_{1,k}, H)$-good coloring of $G$, if $n_1$ is not an enforcer vertex then $u_1$ and $u_2$ must be enforcer vertices.
        \item There exists a $(K_{1,k}, H)$-good coloring of $G$ where $u_1$ and $u_2$ are not enforcer vertices.
        \item There exists a $(K_{1,k}, H)$-good coloring of G where $n_1$ is not an enforcer vertex.
    \end{enumerate}
\end{itemize}

\subsection{$\pmb {(F, H)}$ pairs}
After a careful exploration of viable candidates for $F$ and $H$, we chose all pairs $(F, H)$ with the following properties:
\begin{itemize}
    \item $F$ is star on three or four vertices $(K_{1,2} = P_3$, and $ K_{1,3})$
    \item $H$ is a connected graph on three or four vertices $(P_3, K_3, K_{1,3}, P_4, TK_3, C_4, K_4 - e $, and $K_4)$
    \item $F \not= H$
\end{itemize}
These pairs were chosen because:
\begin{itemize}
    \item they allow for diverse gadget types; for example, determiners only exist when $F \not= H$, and the hardness gadgets are only well-defined for the case where $F$ is a tree.
    \item they allow for solutions that are reasonably sized; using the methodology described in~\cite{wg2026}, we computed the minimal size solutions for each of our problems and found that the largest graph has 10 vertices.
    \item they are underexplored in the literature; while the case where $F$ is a star and $H$ is an arbitrary graph 
    has been explored in the context of extremal graphs~\cite{ds1},
    in the context of our gadget types it was explored only very recently~\cite{hassan2024,wg2026} and that too only for hardness gadgets.
\end{itemize}

\subsection{Verification of results}

Any $(F, H)$-good coloring of $G$ corresponds to a true assignment of the following formula over the variables $\{r_e ~|~ e \in E(G)\}$, where $r_e$ is true if and only if $e$ is red:
\begin{align*}
    \psi_G =&\bigwedge_{e_1, \ldots, e_{k} \in E(G) \text{ form } F}(\overline{r_{e_1}} \lor \overline{r_{e_2}} \lor \cdots \lor \overline{r_{e_k}}) ~~~\land \\ 
    &\bigwedge_{e_1, \ldots, e_{k} \in E(G) \text{ form } H}(r_{e_1} \lor r_{e_2} \lor \cdots \lor r_{e_k})
\end{align*}

Note how the clauses at the top force every copy of $F$ to have at least one blue edge, and the clauses at the bottom force every copy of $H$ to have at least one red edge.

The correspondence between satisfying assignments of $\phi_G$ and $(F,H)$-good colorings of $G$ allows us to easily verify $(F,H)$-goodness and gadget properties. 
For example, given $G$ and $e \in E(G)$, one can verify that $G$ is a $(F,H)$-red-determiner by checking if $\phi_G$ is satisfiable (i.e., $G$ is $(F,H)$-good) and checking if $\phi_G \land (\overline{r_{e}})$ is unsatisfiable (i.e., there is no good coloring when $e$ is blue). Verification scripts for all gadgets follow a similar structure, except for hardness gadgets where it was easier to first compute all $(F,H)$-good colorings (i.e., all satisfying assignments) and then check properties for each vertex.
These scripts are available with our dataset.

\subsection{Notes on our Dataset}
\label{sec:dataset_notes}

\paragraph{Difficulty Tiers.} Note that determiner gadgets only require a constraint on a single edge, whereas other gadget types require constraints on multiple edges/vertices. For the pairs $(F,H)$ in our dataset, determiners can often be constructed by saturating an edge with many copies of $F$ and $H$. As such, we categorize the 26
problems involving determiners as ``easy'' and the other 44 problems in our dataset as ``hard.''

\paragraph{Exclusion of some Problems.} In our dataset, we do not include the construction problems where we are asked to construct hardness gadgets for $(F, H)$ pairs where computing $(F,H)$-goodness is solvable in polynomial-time, since the existence of both of these gadgets would imply P $=$ NP. There are four such cases\footnote{These cases are $(P_3, K_3)$, $(P_3, K_{1,3})$, $(P_3, P_4)$, and $(P_3, TK_3)$.}, giving our dataset a total of $13 \times 6 + 9 \times 4 = 70$ problems.

\paragraph{Minimal Examples.} For each problem in our dataset, we used the methodology and code provided in~\cite{wg2026} to compute the smallest solutions (fewest number of nodes).
This involves iteratively generating $(F,H)$-good graphs up to 10 vertices, removing redundant edges (i.e., edges belonging to neither $F$ nor $H$) when appropriate, and searching for gadgets within this generated set.

\begin{figure}[t]
    \centering
    \includegraphics[width=0.99\linewidth]{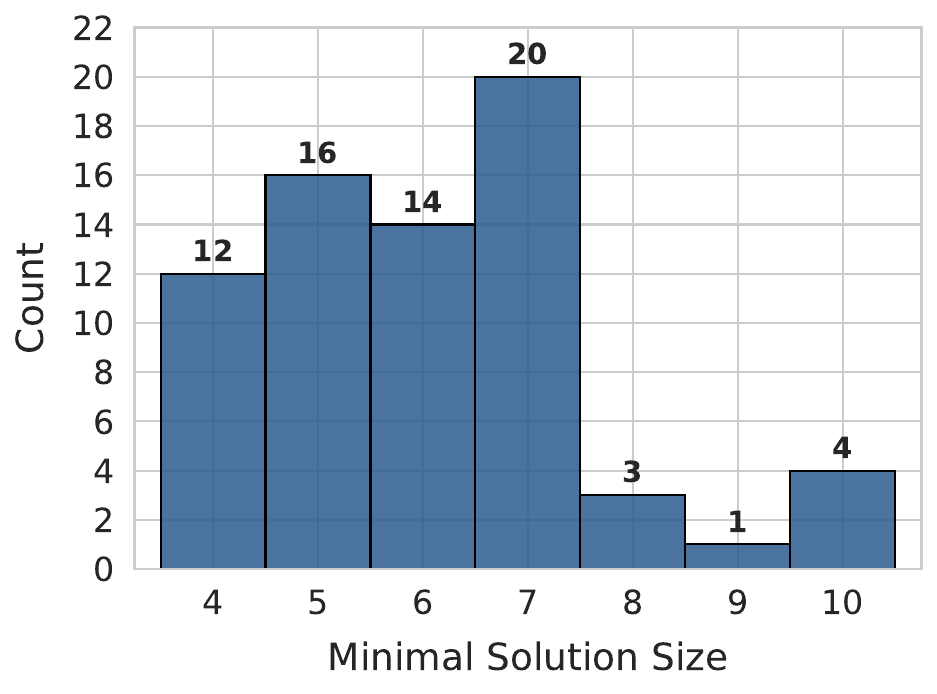}
    \caption{Distribution of minimal solution size.}
    \label{fig:hist_min_sol}
\end{figure}

We provide these minimal examples alongside our dataset. We show the distribution of the sizes of these minimal solutions in Figure~\ref{fig:hist_min_sol}.

\section{Experiments and Results}

We evaluate the performance of five state-of-the-art open-source models with varying sizes: Llama 3.1-8B~\cite{meta_llama31_8b}, Gemma 4-31B~\cite{team2026gemma}, GPT-OSS-120B~\cite{agarwal2025gpt}, Qwen3-235B~\cite{yang2025qwen3}, and DeepSeek-R1~\cite{guo2025deepseek} on \textbf{RamseyGadgets}. All models except Llama 3.1-8B have thinking/reasoning capabilities. We include Llama 3.1-8B in our evaluation because it is comparatively a small and very popular model.
We evaluate these models under three settings: zero-shot, zero-shot with hints, and multiround with access to a graph colorer. We elaborate on each setting and discuss the results below.

\paragraph{Implementation Details.} Our experiments were run on a cluster of 12 nodes where each node had access to one GH200 GPU, 144 Neoverse-V2 CPUs, and 550GB memory. Our code makes use of the following software: vLLM~\cite{kwon2023efficient},
PySAT's~\cite{itk-sat24} implementation of the Glucose SAT solver~\cite{DBLP:journals/ijait/AudemardS18}, NetworkX~\cite{SciPyProceedings_11}, and Grand-Iso~\cite{Matelsky_Motifs_2021}. The hyperparameters for each model and the prompts used in our experiments are provided in the appendix.

\subsection{Experimental Settings}

\begin{table}[ht]
\centering
\small
\setlength{\tabcolsep}{4pt}
\begin{tabular}{l c c c}
\toprule
\textbf{Model} & \textbf{Accuracy (\%)} & \textbf{Pass@5 (\%)} & \textbf{Tokens} \\
\midrule
\multicolumn{4}{c}{\textbf{Zero-shot}} \\
\midrule
Llama-3.1-8B-it & 1.71 $\pm$ 1.86  & 5.71  & 1,175.4  \\
gemma-4-31B-it        & 36.00 $\pm$ 2.35 & 51.43 & 18,118.0 \\
gpt-oss-120b          & 21.14 $\pm$ 1.86 & 38.57 & 13,513.8 \\
Qwen3-235B-A22B       & 26.16 $\pm$ 3.90 & 45.07 & 36,607.1 \\
DeepSeek-R1           & 10.29 $\pm$ 1.20 & 21.43 & 21,537.2 \\
\midrule
\multicolumn{4}{c}{\textbf{Zero-shot + Structural Hint}} \\
\midrule
Llama-3.1-8B-it & 3.43 $\pm$ 0.78  & 5.71  & 1,488.7  \\
gemma-4-31B-it        & 38.57 $\pm$ 5.25 & 54.29 & 18,231.5 \\
gpt-oss-120b          & 18.86 $\pm$ 1.86 & 37.14 & 11,697.6 \\
Qwen3-235B-A22B       & 24.40 $\pm$ 6.34 & 42.25 & 36,676.1 \\
DeepSeek-R1           & 10.00 $\pm$ 2.26 & 18.57 & 21,798.6 \\
\midrule
\multicolumn{4}{c}{\textbf{Zero-shot + Size Hint}} \\
\midrule
Llama-3.1-8B-it & 2.86 $\pm$ 1.01  & 8.57  & 1,155.0  \\
gemma-4-31B-it        & 40.00 $\pm$ 2.86 & 55.71 & 18,209.7 \\
gpt-oss-120b          & 25.71 $\pm$ 5.05 & 48.57 & 13,362.1 \\
Qwen3-235B-A22B       & 27.71 $\pm$ 2.96 & 44.29 & 31,337.4 \\
DeepSeek-R1           & 11.43 $\pm$ 3.19 & 21.43 & 19,266.4 \\
\midrule
\multicolumn{4}{c}{\textbf{Multiround}} \\
\midrule
Llama-3.1-8B-it & 1.69 $\pm$ 1.18  & 8.45  & 13,807.3 \\
gemma-4-31B-it        & 51.83 $\pm$ 3.05 & 66.20 & 56,124.9 \\
gpt-oss-120b          & 25.07 $\pm$ 5.02 & 42.25 & 38,167.1 \\
Qwen3-235B-A22B       & 28.45 $\pm$ 10.13& 54.93 & 69,706.5 \\
DeepSeek-R1           & 11.92 $\pm$ 2.32 & 25.35 & 49,755.2 \\
\bottomrule
\end{tabular}
\caption{Model performance (mean accuracy $\pm$ standard deviation \%, Pass@5 \%) and average total tokens across evaluation settings.}
\label{tab:experiment_results}
\end{table}

\subsubsection{Zero-shot.}
In this setting, the model is given all required definitions and asked to generate a graph adhering to the given constraints.
The model has one round to provide an answer after reasoning. 

\subsubsection{Zero-shot with Hints.}
This setting is similar to the zero-shot setting, but we also provide the model a hint relevant to its task. We use three different hints:

\begin{itemize}
    \item \textbf{Structural.} In this setting, we tell the model that in any $(F,H)$-good coloring of a graph $G$, any edge that does not belong to $H$ can always be colored blue. This fact, albeit simple, is used to prune the search space for computational approaches~\cite{wg2026}.
    \item \textbf{Size.} In this setting, we tell the model that the desired graph is known to exist on $n$ vertices, where $n$ is the size of the minimal example we computed. 
This setting mimics the scenario when we know an object exists due to a
nonconstructive proof but don't know exactly what the object is.
    \item \textbf{Relevant Example.} In this setting, we give the model an example of a similar problem whose solution could be generalized to the target problem. Particularly, for prompts concerning $(K_{1,3}, H)$-goodness problems, we give the model an example of a similar gadget for the $(P_3, H)$-goodness setting if it exists. This gives us a total of 30 problems. This setting is inspired by the fact that constructions involving similar $F$ or $H$ can often be generalized, as seen in~\cite{hassan2024} and~\cite{wg2026}.
\end{itemize}

\subsubsection{Multiround with Access to Coloring Tool.}
In this setting, we provide the model access to a graph colorer that returns all $(F,H)$-good colorings of a given graph for the relevant problem. The model has a total of three rounds to complete the task: two rounds for exploration, and one round to aggregate its findings and provide an answer.

\subsection{Results}

We repeat each experiment five times and report: (1) the average accuracy and standard deviation over each run, (2) the
Pass@5 percentage (i.e., the percentage of problems that were solved correctly at least once across the five runs), and (3) the average number of tokens used. Our results (Table~\ref{tab:experiment_results}) show that Gemma-4-31B is the best-performing model across all settings based on mean accuracy and Pass@5, followed by Qwen-235B, gpt-oss-120b, and DeepSeek-R1, while Llama-3.1-8B is the worst-performing one. The low performance of Llama-3.1-8B is expected as it is not a thinking model. Gemma-4-31B achieves a mean accuracy of 36.00 $\pm$ 2.35\% and Pass@5 of 51.43\% in the Zero-shot setting. 
The best results are achieved by Gemma-31B in the Multiround setting, where it has access to a graph colorer: it achieves a mean accuracy of 51.83 $\pm$ 3.05\% and Pass@5 of 66.20\%. We observe in our analysis that a majority of this success comes from the ``easy'' difficulty tier of our dataset.
Our results demonstrate that these popular LLMs struggle to achieve high performance on \textbf{RamseyGadgets}.

\begin{figure}[t]
    \centering
    \includegraphics[width=0.99\linewidth]{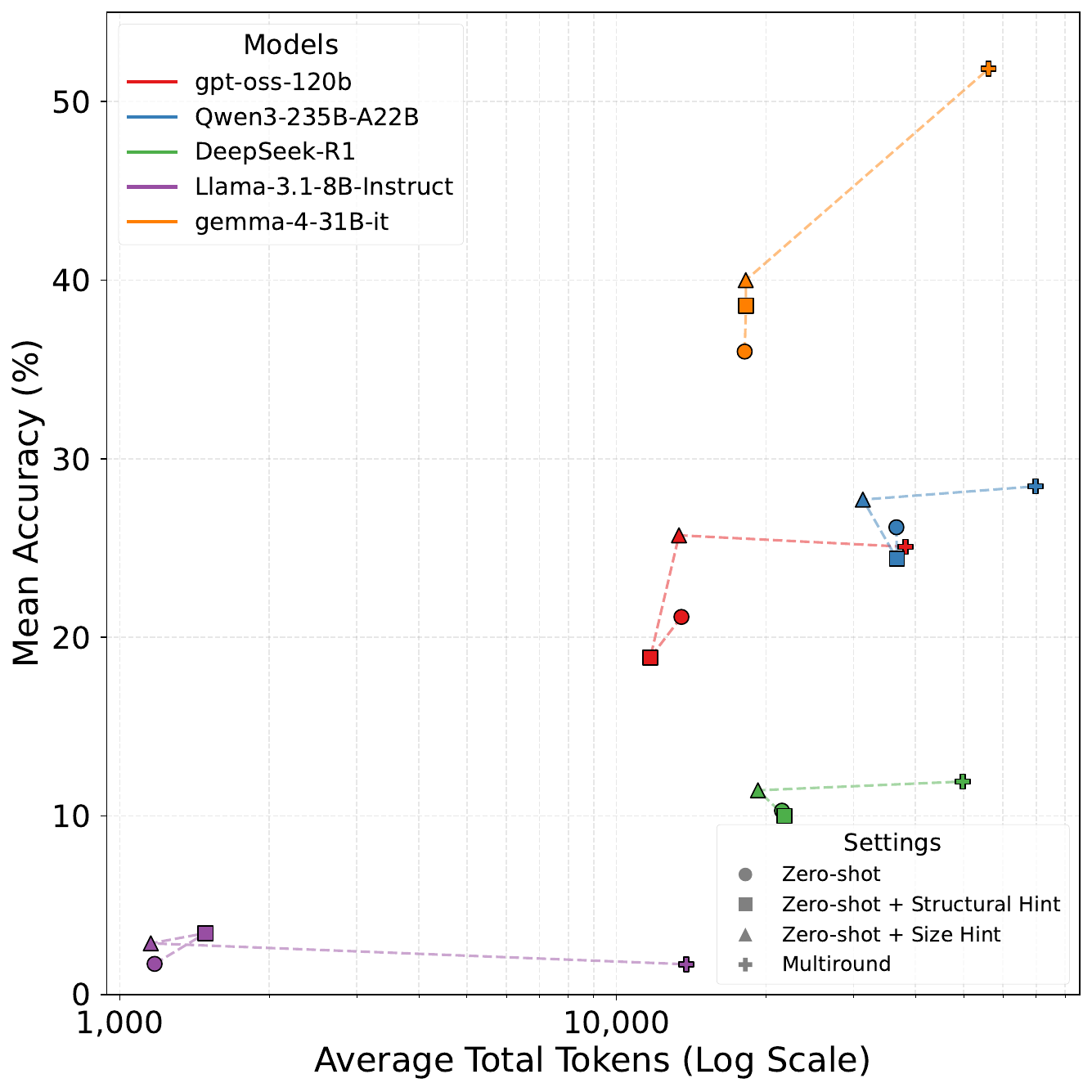}
    \caption{Accuracy vs. Token Cost Trade-off across Models \& Settings}
    \label{fig:acc_vs_tokens}
\end{figure}

\subsubsection{Computational Cost.}
Table~\ref{tab:experiment_results} also shows the average total tokens consumed by the models. The Qwen-235B model turns out to be the most expensive in terms of token consumption while achieving the second-best performance. Figure~\ref{fig:acc_vs_tokens} shows the mean accuracy versus average total token cost trade-off across all models and settings. Gemma-4-31B turns out to offer the best trade-off between computational cost and performance.

\subsubsection{Impact of Structural and Size Hints.}

Table~\ref{tab:experiment_results} shows that 
our structural hint typically had a negative effect, with most models showing a slight decrease in performance as compared to the zero-shot setting. The only exception to this is Gemma-4-31B, where the accuracy and Pass@5 improve.  
On the other hand, the size hint generally improves the performance of all models, making a substantial (10\%) impact on the results for gpt-oss-120b.

\begin{table}[t]
\centering
\small
\setlength{\tabcolsep}{4pt}
\begin{tabular}{lccc}
\toprule
\textbf{Model} & \textbf{Accuracy (\%)} & \textbf{Pass@5 (\%)} & \textbf{Tokens} \\
\midrule
\multicolumn{4}{c}{\textbf{Zero-shot}} \\
\midrule
Llama-3.1-8B-it & 1.33 $\pm$ 1.83 & 1.33 & 1,181.3 \\
gemma-4-31B-it & 20.00 $\pm$ 2.36 & 20.00 & 19,343.0 \\
gpt-oss-120b & 10.00 $\pm$ 4.08 & 10.00 & 11,772.5 \\
Qwen3-235B-A22B & 13.67 $\pm$ 4.15 & 13.57 & 41,503.3 \\
DeepSeek-R1 & 6.67 $\pm$ 2.36 & 6.67 & 22,363.3 \\
\midrule
\multicolumn{4}{c}{\textbf{Zero-shot + Example Hint}} \\
\midrule
Llama-3.1-8B-it & 3.33 $\pm$ 0.00 & 3.33 & 2,299.3 \\
gemma-4-31B-it & 25.33 $\pm$ 3.80 & 36.67 & 18,768.9 \\
gpt-oss-120b & 11.33 $\pm$ 3.80 & 20.00 & 14,407.5 \\
Qwen3-235B-A22B & 14.19 $\pm$ 1.44 & 19.35 & 38,195.0 \\
DeepSeek-R1 & 4.67 $\pm$ 1.83 & 6.67 & 22,133.1 \\
\bottomrule
\end{tabular}
\caption{Model performance (mean accuracy $\pm$ standard deviation \%, Pass@5 \%) and average total tokens for the cases considered in the Zero-shot + Example Hint setting.}
\label{tab:results_example}
\end{table}

\subsubsection{Impact of Example Hint.}
Table~\ref{tab:results_example} shows the comparative performance of the models in the Zero-shot setting against the Zero-shot + Example Hint setting. Note that in this case, we are only evaluating over the 30 problems to which this setting applies. We observe that providing an example substantially improves the performance of all models, with the exception of DeepSeek-R1.

\begin{figure}[ht]
  \centering
  \begin{subfigure}[b]{0.69\linewidth}
    \centering
    \includegraphics[width=\linewidth]{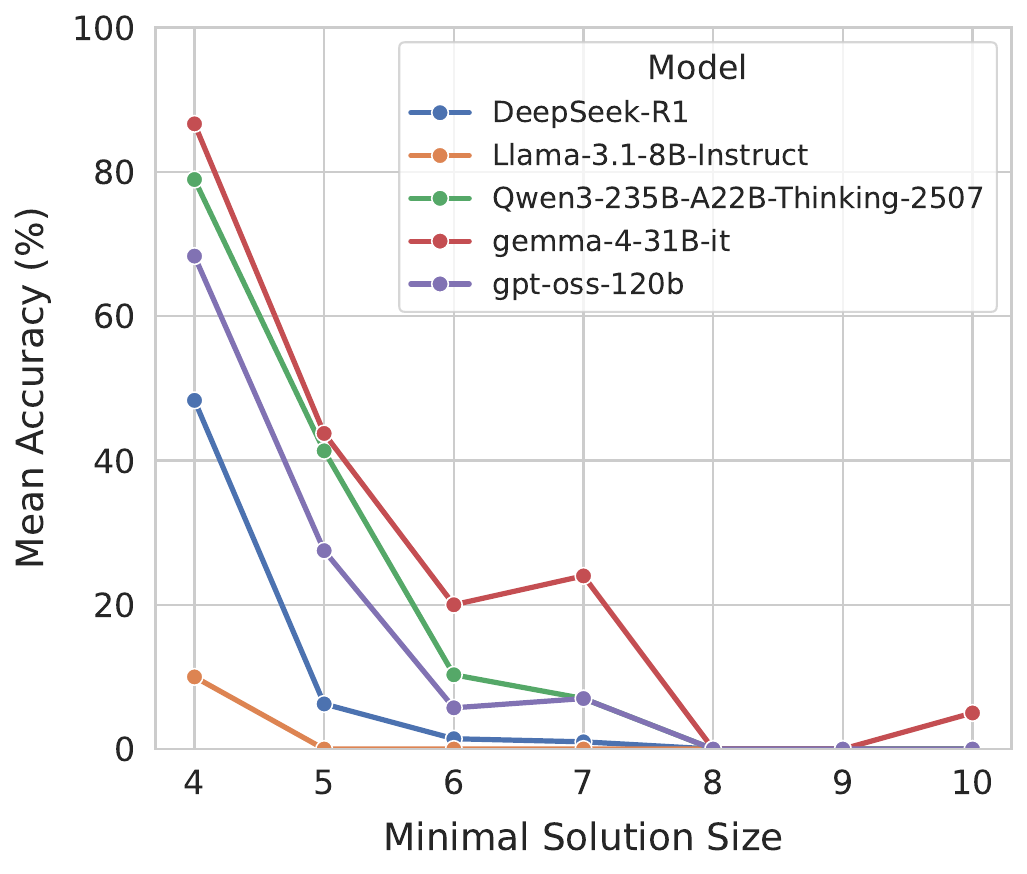}
    \caption{Zero-shot}
    \label{fig:mean_accuracy_zeroshot}
  \end{subfigure} \hfill
  \begin{subfigure}[b]{0.69\linewidth}
    \centering
    \includegraphics[width=\linewidth]{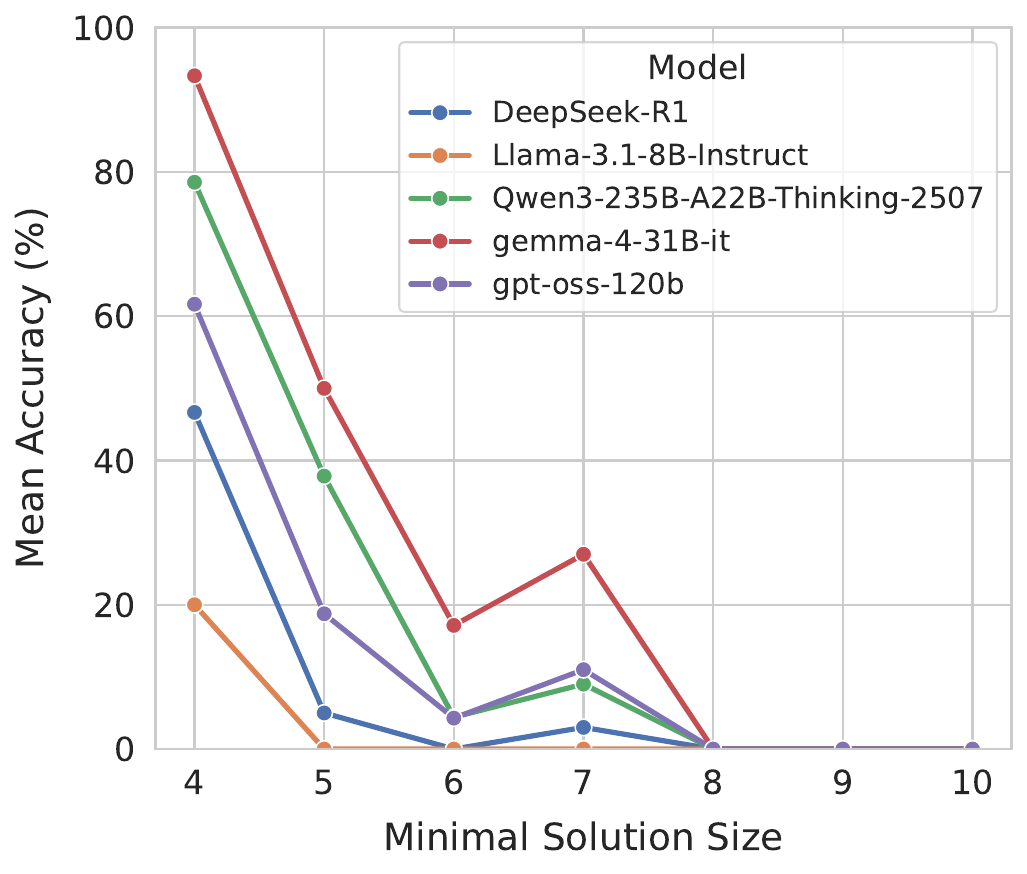}
    \caption{Zero-shot + Structural Hint}
    \label{fig:mean_accuracy_zeroshot_struct}
  \end{subfigure} \hfill
  \begin{subfigure}[b]{0.69\linewidth}
    \centering
    \includegraphics[width=\linewidth]{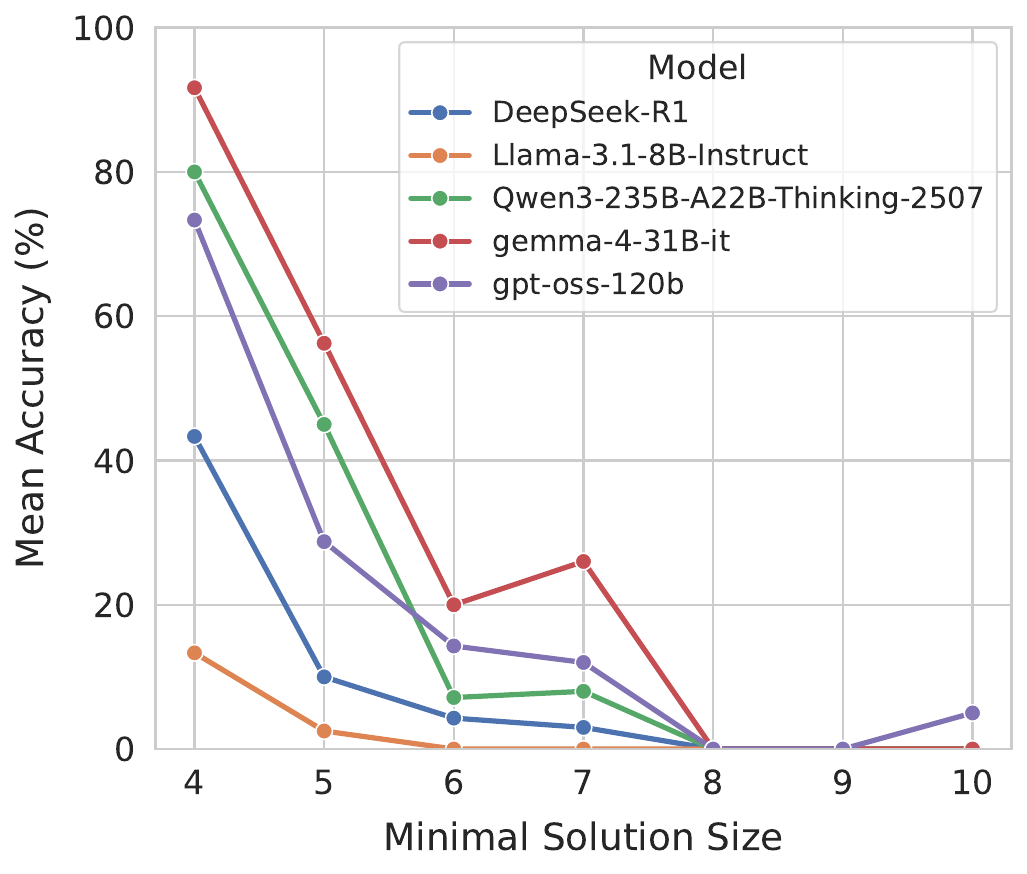}
    \caption{Zero-shot + Size Hint}
    \label{fig:mean_accuracy_zeroshot_size}
  \end{subfigure} \hfill
  \begin{subfigure}[b]{0.69\linewidth}
    \centering
    \includegraphics[width=\linewidth]{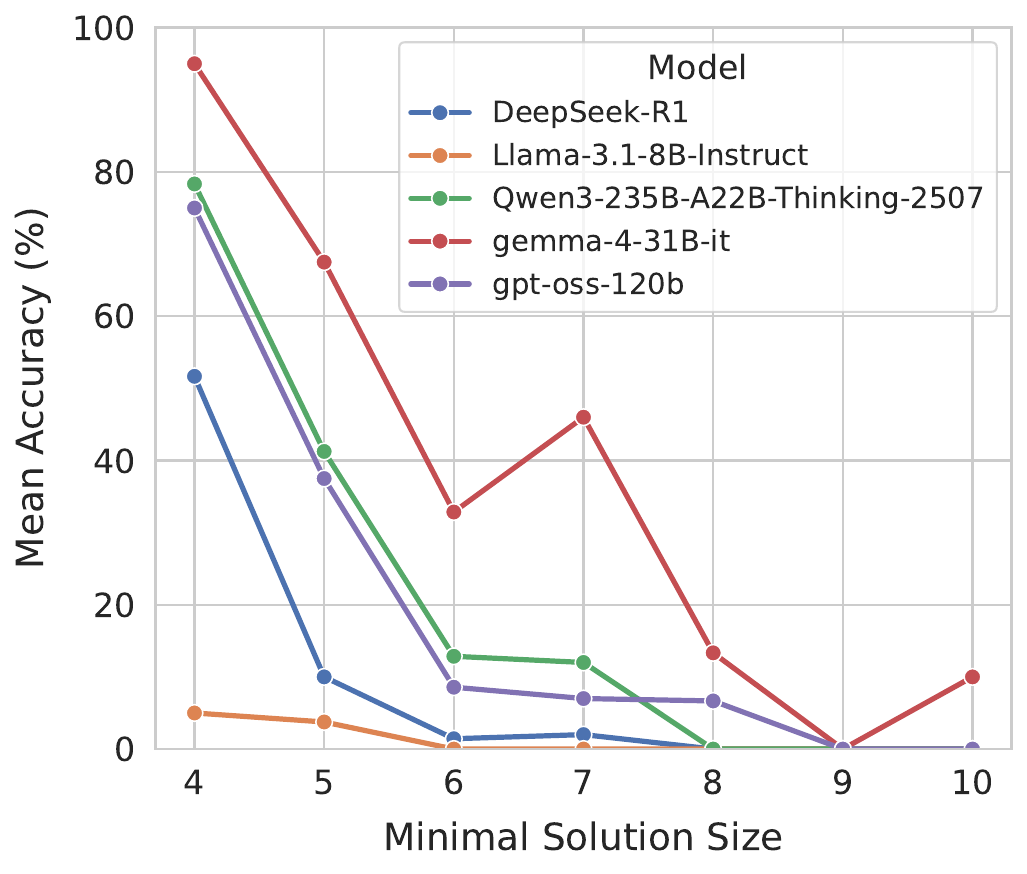}
    \caption{Multiround}
    \label{fig:mean_accuracy_multiround}
  \end{subfigure}  
  \caption{Mean Accuracy vs.\ minimal solution size.}
  \label{fig:mean_accuracy_min_ex_size}
\end{figure}

\subsubsection{Impact of Coloring Tool.}
Having access to a coloring tool in the multiround setting substantially improves the performance of all models. Notably, this setting sees the best results on the dataset; Gemma-4-31B achieves a mean accuracy and Pass@5 of 51.83\% $\pm$ 3.05 and 66.20\%.
However, we also observe that token usage increases substantially, with an average of 2.5x more tokens across all models when compared to the Zero-shot setting.

\subsubsection{Impact of Minimal Solution Size.}
Figure~\ref{fig:mean_accuracy_min_ex_size} shows the graphs for mean accuracy against the minimal solution size across all models and settings. 
We notice that the performance degrades as the number of vertices in the minimal solution increases. This demonstrates a very interesting pattern highlighting the increasing difficulty of the problems based on a minimal solution.
We note that our dataset only has 8 problems that have minimal solutions with at least 8 vertices (see Figure~\ref{fig:hist_min_sol}), making it difficult to assert this trend. However, this trend holds even across problems with minimal solution sizes 4 to 7, which are well-represented in our dataset.
We also note that the problems requiring a solution on at least 10 vertices solved by the models are part of the easy difficulty tier of our dataset; these are $(K_{1,3}, K_4)$-red- and $(K_{1,3}, K_4)$-blue-determiners.

\subsubsection{Results across Difficulty Tiers.}
\begin{figure}[ht]
  \centering
  \begin{subfigure}[b]{0.77\linewidth}
    \centering
    \includegraphics[width=\linewidth]{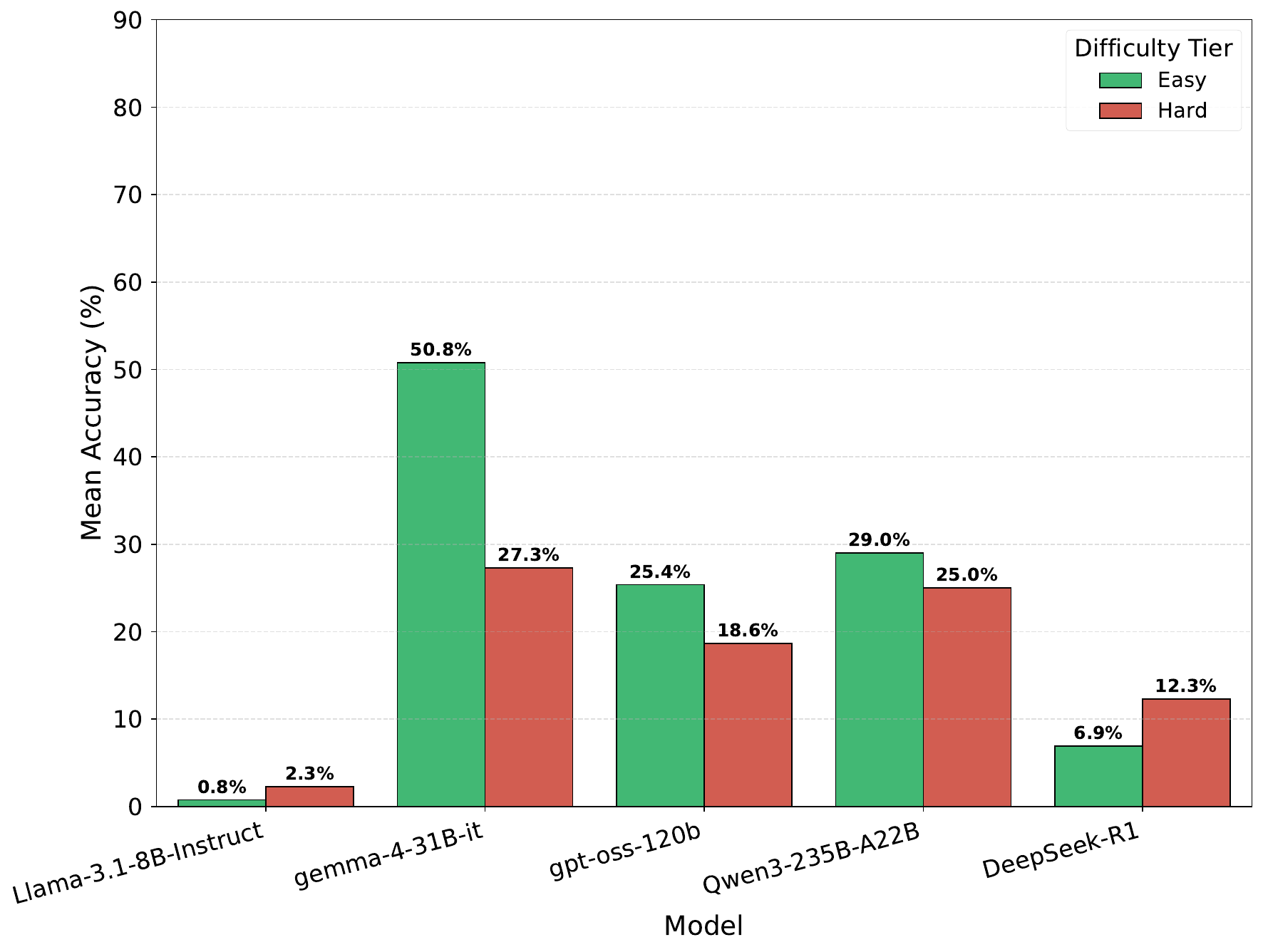}
    \caption{Zero-shot}
    \label{fig:difficulty_zeroshot}
  \end{subfigure} \hfill
  \begin{subfigure}[b]{0.77\linewidth}
    \centering
    \includegraphics[width=\linewidth]{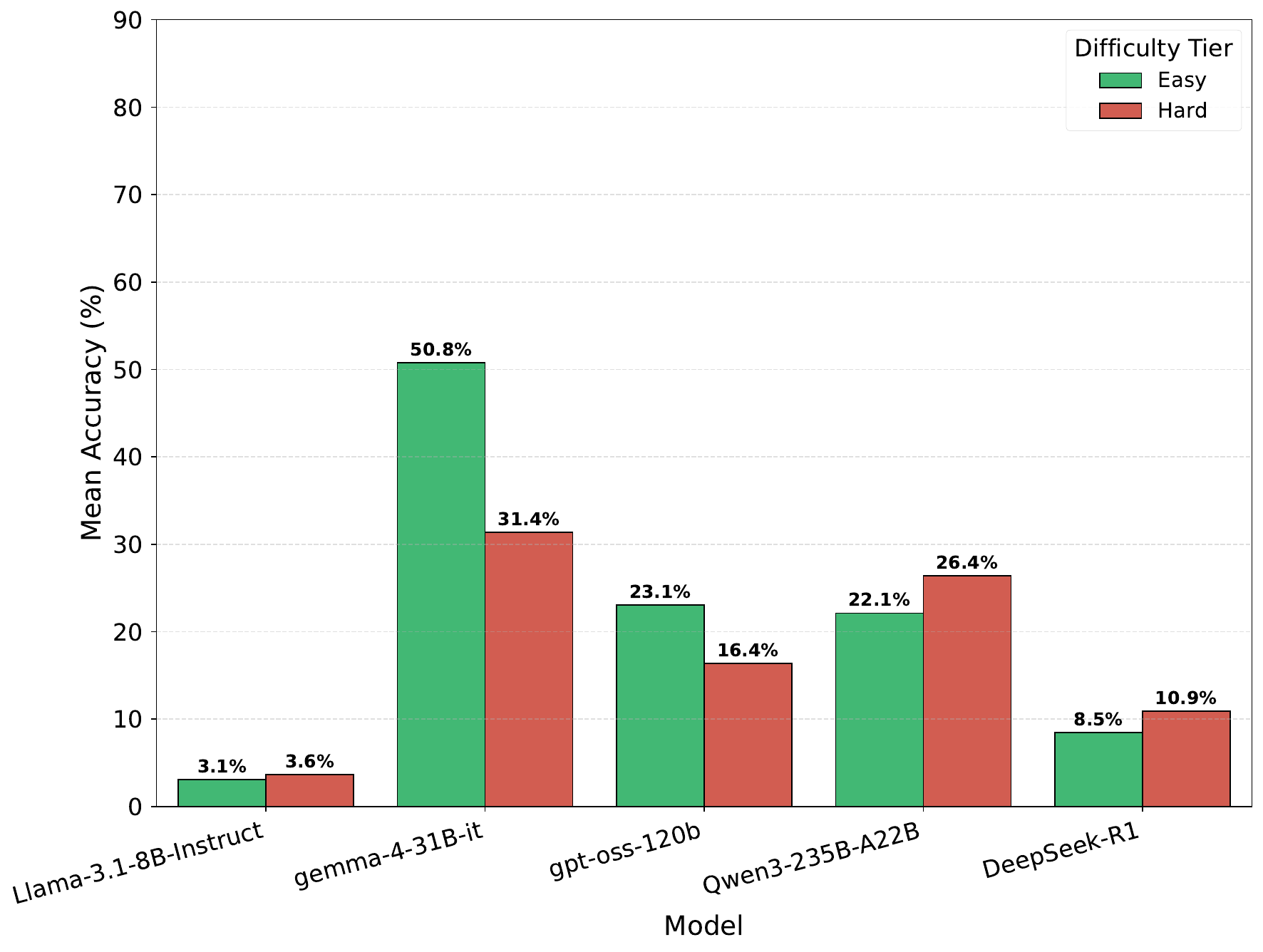}
    \caption{Zero-shot + Structural Hint}
    \label{fig:difficulty_zeroshot_struct}
  \end{subfigure} \hfill
  \begin{subfigure}[b]{0.77\linewidth}
    \centering
    \includegraphics[width=\linewidth]{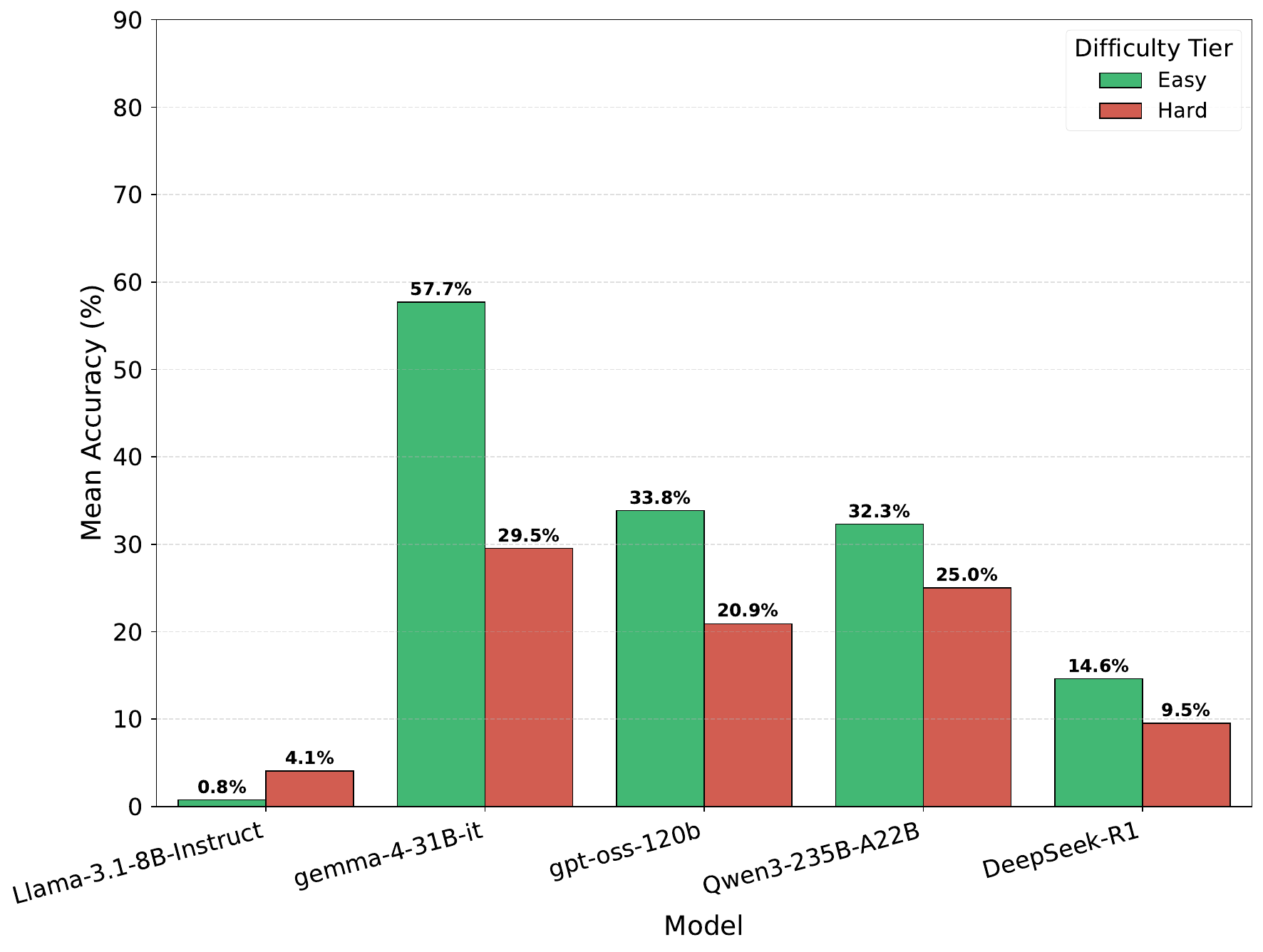}
    \caption{Zero-shot + Size Hint}
    \label{fig:difficulty_zeroshot_size}
  \end{subfigure} \hfill
  \begin{subfigure}[b]{0.77\linewidth}
    \centering
    \includegraphics[width=\linewidth]{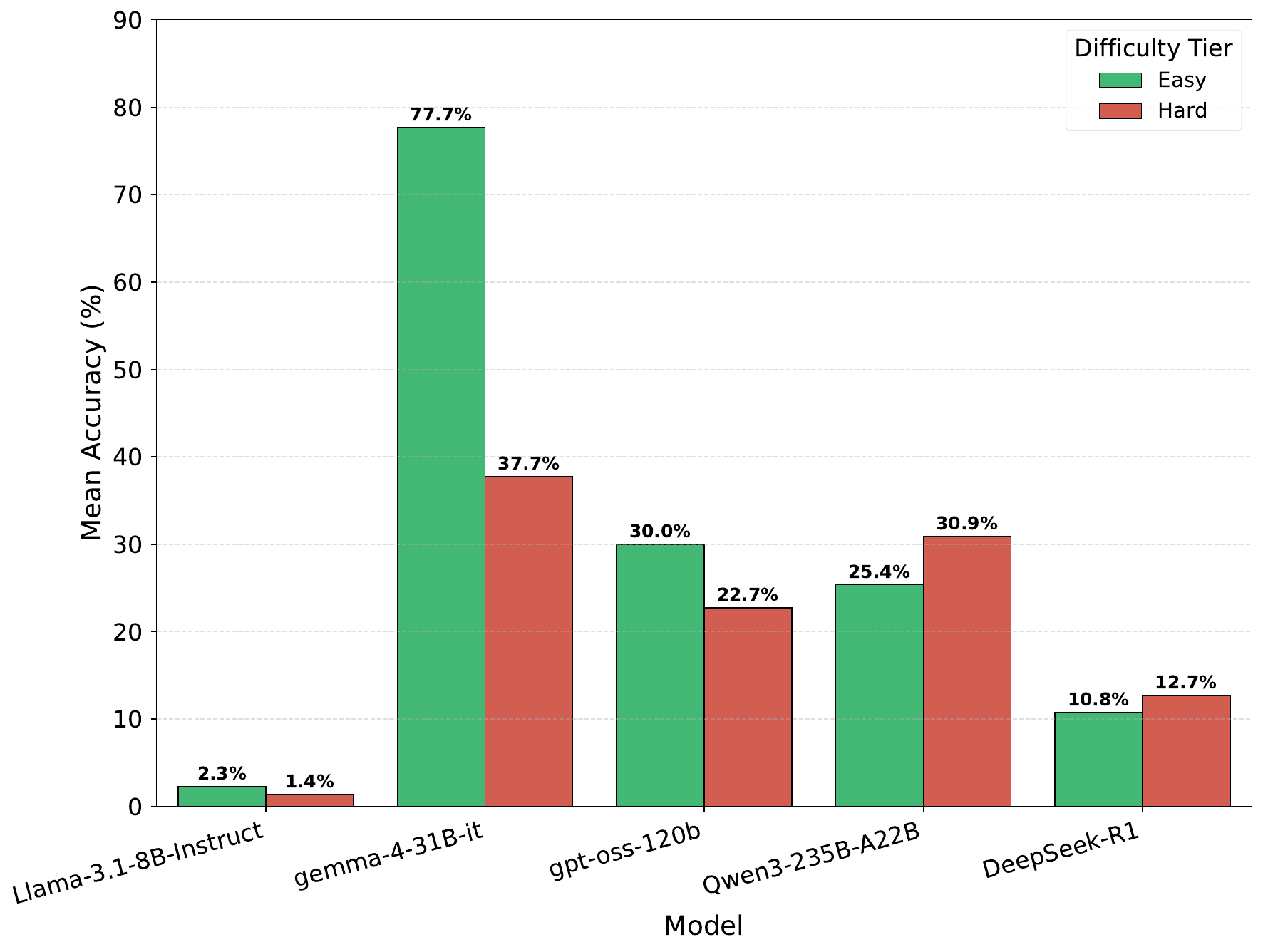}
    \caption{Multiround}
    \label{fig:difficulty_multiround}
  \end{subfigure} 
  \caption{Model Accuracy by difficulty.}
  \label{fig:difficulty_breakdown}
\end{figure}

Figure~\ref{fig:difficulty_breakdown} shows the performance of all models across settings based on the difficulty tier (``easy'' and ``hard'') of the problems. We note that the models struggle on the hard problems in our dataset, with the highest mean accuracy being 37.7\% for
Gemma4-31B.

\section{Discussion}

In this section, we discuss our results in the context of the research questions posed in the introduction.

\begin{itemize}
    \item \textbf{RQ1.} \textit{Does providing a domain-specific hint help LLMs perform better at the task?}
    Our results demonstrate that hints can improve the performance of LLMs if the hint is strong enough. Conversely, the performance can also be stymied if the hint is too general. As observed in our results, providing a general structural hint can negatively impact performance, while the more specific size and example hints almost always improved performance.
    \item \textbf{RQ2.} \textit{Does performance decrease as minimal solution size grows?}
    Our results show that the performance substantially decreases as minimal solution size grows, even when going from 4 vertices to 7 vertices.
    \item \textbf{RQ3.} \textit{Are LLMs able to perform better when given access to tools specialized for the task?}
    Our results show a substantial performance improvement when models are given a tool that computes $(F, H)$-good colorings. 
\end{itemize}

\section{Conclusion}

In this work we presented \textbf{RamseyGadgets}, a novel specialized dataset of 70 graph construction problems that are less examined and have reasonably sized and easily verifiable solutions.
These problems are based on finding Ramsey-good graphs with special colorings. 
To avoid saturation, \textbf{RamseyGadgets} can easily be expanded to include more problems; each problem in the dataset is based on generating $(F, H)$-good graphs, and simply varying $F$ and $H$ will give us new problems. The limitation of this approach is that as $F$ and $H$ grow larger, it becomes increasingly difficult to compute minimal solutions. However, this limitation only affects the scenario where one wants to provide this solution or its size as a hint. Other ways of obtaining new problems include avoiding multiple subgraphs in both colors and using more than two colors. 

Our evaluation of five popular LLMs on \textbf{RamseyGadgets} shows that these models struggle to reliably solve graph construction problems, despite that fact that the largest minimal solution size to a problem is 10 vertices. 
We demonstrated how our specialized dataset allows us to ask and analyze research questions about LLMs specific to graph construction, such as the impact of hints and tools.

We see our work as an important first step towards improving the performance of LLMs on graph construction problems, allowing researchers to delegate these arduous yet frequently-encountered tasks to these models.

\bibliography{aaai2027}

@book{DBLP:books/daglib/0023084,
  author       = {S. Arora and
                  B. Barak},
  title        = {Computational Complexity - {A} Modern Approach},
  publisher    = {Cambridge University Press},
  year         = {2009},
}

@article{ds1,
author = {S. Radziszowski},
journal = {Electronic Journal of Combinatorics},
volume = {DS1},
month = {April},
pages = {1--149},
booktitle = {{Ramsey Theory: Yesterday, Today, and Tomorrow}},
title = {{Small Ramsey Numbers}},
year = {2026},
url = {https://www.combinatorics.org/},
}

@book{DBLP:books/fm/GareyJ79,
  author       = {M. R. Garey and
                  D. S. Johnson},
  title        = {Computers and Intractability: {A} Guide to the Theory of {NP}-Completeness},
  publisher    = {W. H. Freeman},
  year         = {1979},
  isbn         = {0-7167-1044-7},
}

@inproceedings{wg2026,
  author       = {Zohair Raza Hassan},
  editor       = {Jan Goedgebeur and
                  Pawel Rzazewski},
  title        = {The Complexity of Ramsey Arrowing: {A} Computational Approach for
                  Hardness Proofs},
  booktitle    = {52nd International Workshop on Graph-Theoretic Concepts in Computer
                  Science, {WG} 2026, Kortrijk, Belgium, June 2-4, 2026},
  series       = {LIPIcs},
  volume       = {376},
  pages        = {25:1--25:20},
  publisher    = {Schloss Dagstuhl - Leibniz-Zentrum f{\"{u}}r Informatik},
  year         = {2026},
  url          = {https://doi.org/10.4230/LIPIcs.WG.2026.25},
  doi          = {10.4230/LIPICS.WG.2026.25},
  bibsource    = {dblp computer science bibliography, https://dblp.org}
}

@article{DBLP:journals/corr/abs-2603-09172,
  author       = {A. Nagda and
                  P. Raghavan and
                  A. Thakurta},
  title        = {Reinforced Generation of Combinatorial Structures: Ramsey Numbers},
  journal      = {CoRR},
  volume       = {abs/2603.09172},
  year         = {2026},
}

@article{DBLP:journals/corr/abs-2506-13131,
  author       = {A. Novikov and
                  N. and
                  M. Eisenberger and
                  E. Dupont and
                  others},
  title        = {AlphaEvolve: {A} coding agent for scientific and algorithmic discovery},
  journal      = {CoRR},
  volume       = {abs/2506.13131},
  year         = {2025},
  url          = {https://doi.org/10.48550/arXiv.2506.13131},
  doi          = {10.48550/ARXIV.2506.13131},
  eprinttype   = {arXiv},
  eprint       = {2506.13131},
  bibsource    = {dblp computer science bibliography, https://dblp.org}
}

@article{petersen1898sur,
  author  = {Petersen, J.},
  title   = {Sur le théorème de Tait},
  journal = {L'Intermédiaire des Mathématiciens},
  volume  = {5},
  pages   = {225--227},
  year    = {1898}
}

@article{schlafli1858attempt,
  title={An Attempt to Determine the Twenty-seven Lines upon a Surface of the Third Order and to Divide such Surfaces into Species in Reference to the Reality of the Lines upon the Surface},
  author={L. Schl{\"a}fli},
  journal={The Quarterly Journal of Pure and Applied Mathematics},
  volume={2},
  pages={110--120},
  year={1858}
}

@article{DBLP:journals/dam/GoedgebeurO22,
  author       = {Goedgebeur, J. and
                  Van Overberghe, S.},
  title        = {{New bounds for {R}amsey numbers $R(K_k-e, K_l-e)$}},
  journal      = {Discrete Applied Mathematics},
  volume       = {307},
  pages        = {212--221},
  year         = {2022},
  
}

@inproceedings{DBLP:conf/soda/InoueKMMS26,
  author       = {Yuta Inoue and
                  Ken{-}ichi Kawarabayashi and
                  Atsuyuki Miyashita and
                  Bojan Mohar and
                  Tomohiro Sonobe},
  editor       = {Kasper Green Larsen and
                  Barna Saha},
  title        = {Three-edge-coloring (Tait coloring) cubic graphs and nowhere-zero
                  4-flow for graphs on the torus},
  booktitle    = {Proceedings of the 2026 Annual {ACM-SIAM} Symposium on Discrete Algorithms,
                  {SODA} 2026, Vancouver, BC, Canada, January 11-14, 2026},
  pages        = {6133--6165},
  publisher    = {{SIAM}},
  year         = {2026},
  url          = {https://doi.org/10.1137/1.9781611978971.218},
  doi          = {10.1137/1.9781611978971.218},
  bibsource    = {dblp computer science bibliography, https://dblp.org}
}

@inproceedings{DBLP:conf/icml/BalunovicD0PV25,
  author       = {Mislav Balunovic and
                  Jasper Dekoninck and
                  Nikola Jovanovic and
                  Ivo Petrov and
                  Martin T. Vechev},
  editor       = {Aarti Singh and
                  Maryam Fazel and
                  Daniel Hsu and
                  Simon Lacoste{-}Julien and
                  Felix Berkenkamp and
                  Tegan Maharaj and
                  Kiri Wagstaff and
                  Jerry Zhu},
  title        = {MathConstruct: Challenging {LLM} Reasoning with Constructive Proofs},
  booktitle    = {Forty-second International Conference on Machine Learning, {ICML}
                  2025, Vancouver, BC, Canada, July 13-19, 2025},
  series       = {Proceedings of Machine Learning Research},
  volume       = {267},
  publisher    = {{PMLR} / OpenReview.net},
  year         = {2025},
  url          = {https://proceedings.mlr.press/v267/balunovic25a.html},
  bibsource    = {dblp computer science bibliography, https://dblp.org}
}

@article{Scha,
  author       = {Marcus Schaefer},
  title        = {Graph Ramsey Theory and the Polynomial Hierarchy},
  journal      = {Journal of Computer and System Sciences},
  volume       = {62},
  number       = {2},
  pages        = {290--322},
  year         = {2001},
  url          = {https://doi.org/10.1006/jcss.2000.1729},
  doi          = {10.1006/JCSS.2000.1729},
  bibsource    = {dblp computer science bibliography, https://dblp.org}
}

@article{burr1976graphs,
  title={{On Graphs of {R}amsey Type}},
  author={S. A. Burr and P. Erd\H{o}s and L. Lov{\'a}sz},
  journal={Ars Combinatoria},
  volume={1},
  number={1},
  pages={167--190},
  year={1976}
}

@article{Bu3,
  author = {S. A. Burr},
  title = {{On the Computational Complexity of Ramsey-Type Problems}},
  journal = {Mathematics of Ramsey Theory},
  year = {1990},
  volume = {5},
  pages = {46-52}
}

@inproceedings{hassan2024,
  author       = {Zohair Raza Hassan},
  editor       = {Rastislav Kr{\'{a}}lovic and
                  Anton{\'{\i}}n Kucera},
  title        = {The Complexity of $(P_3,H)$-Arrowing and Beyond},
  booktitle    = {49th International Symposium on Mathematical Foundations of Computer
                  Science, {MFCS} 2024, Bratislava, Slovakia, August 26-30, 2024},
  series       = {LIPIcs},
  volume       = {306},
  pages        = {59:1--59:16},
  publisher    = {Schloss Dagstuhl - Leibniz-Zentrum f{\"{u}}r Informatik},
  year         = {2024},
  url          = {https://doi.org/10.4230/LIPIcs.MFCS.2024.59},
  doi          = {10.4230/LIPICS.MFCS.2024.59},
  bibsource    = {dblp computer science bibliography, https://dblp.org}
}

@inproceedings{itk-sat24,
  author       = {Alexey Ignatiev and
                  Zi Li Tan and
                  Christos Karamanos},
  editor       = {Supratik Chakraborty and
                  Jie{-}Hong Roland Jiang},
  title        = {Towards Universally Accessible {SAT} Technology},
  booktitle    = {27th International Conference on Theory and Applications of Satisfiability
                  Testing, {SAT} 2024, Pune, India, August 21-24, 2024},
  series       = {LIPIcs},
  volume       = {305},
  pages        = {16:1--16:11},
  publisher    = {Schloss Dagstuhl - Leibniz-Zentrum f{\"{u}}r Informatik},
  year         = {2024},
  url          = {https://doi.org/10.4230/LIPIcs.SAT.2024.16},
  doi          = {10.4230/LIPICS.SAT.2024.16},
  bibsource    = {dblp computer science bibliography, https://dblp.org}
}

@article{DBLP:journals/ijait/AudemardS18,
  author       = {Gilles Audemard and
                  Laurent Simon},
  title        = {On the Glucose {SAT} Solver},
  journal      = {Int. J. Artif. Intell. Tools},
  volume       = {27},
  number       = {1},
  pages        = {1840001:1--1840001:25},
  year         = {2018},
  url          = {https://doi.org/10.1142/S0218213018400018},
  doi          = {10.1142/S0218213018400018},
  bibsource    = {dblp computer science bibliography, https://dblp.org}
}

@InProceedings{SciPyProceedings_11,
  author =       {Aric A. Hagberg and Daniel A. Schult and Pieter J. Swart},
  title =        {Exploring Network Structure, Dynamics, and Function using NetworkX},
  booktitle =   {Proceedings of the 7th Python in Science Conference},
  pages =     {11 - 15},
  address = {Pasadena, CA USA},
  year =      {2008},
  editor =    {Ga\"el Varoquaux and Travis Vaught and Jarrod Millman},
}

@article{Matelsky_Motifs_2021, 
    title={{DotMotif: an open-source tool for connectome subgraph isomorphism search and graph queries}},
    volume={11}, 
    number={1}, 
    journal={Scientific Reports}, 
    publisher={Springer}, 
    author={Matelsky, J. K. and Reilly, E. P. and Johnson, E. C. and Stiso, J. and Bassett, D. S. and Wester, B. A. and Gray-Roncal, W.},
    year={2021}, 
    month={Jun}
}

@inproceedings{kwon2023efficient,
  author       = {Woosuk Kwon and
                  Zhuohan Li and
                  Siyuan Zhuang and
                  Ying Sheng and
                  Lianmin Zheng and
                  Cody Hao Yu and
                  Joseph Gonzalez and
                  Hao Zhang and
                  Ion Stoica},
  editor       = {Jason Flinn and
                  Margo I. Seltzer and
                  Peter Druschel and
                  Antoine Kaufmann and
                  Jonathan Mace},
  title        = {Efficient Memory Management for Large Language Model Serving with
                  PagedAttention},
  booktitle    = {Proceedings of the 29th Symposium on Operating Systems Principles,
                  {SOSP} 2023, Koblenz, Germany, October 23-26, 2023},
  pages        = {611--626},
  publisher    = {{ACM}},
  year         = {2023},
  url          = {https://doi.org/10.1145/3600006.3613165},
  doi          = {10.1145/3600006.3613165},
  bibsource    = {dblp computer science bibliography, https://dblp.org}
}

@article{jin2024large,
  title={Large language models on graphs: A comprehensive survey},
  author={Jin, Bowen and Liu, Gang and Han, Chi and Jiang, Meng and Ji, Heng and Han, Jiawei},
  journal={IEEE Transactions on Knowledge and Data Engineering},
  volume={36},
  number={12},
  pages={8622--8642},
  year={2024},
  publisher={IEEE}
}

@article{wang2025graph,
  title={Graph machine learning in the era of large language models (llms)},
  author={Wang, Shijie and Huang, Jiani and Chen, Zhikai and Song, Yu and Tang, Wenzhuo and Mao, Haitao and Fan, Wenqi and Liu, Hui and Liu, Xiaorui and Yin, Dawei and others},
  journal={ACM Transactions on Intelligent Systems and Technology},
  volume={16},
  number={5},
  pages={1--40},
  year={2025},
  publisher={ACM New York, NY}
}

@inproceedings{fan-etal-2024-nphardeval,
    title = "{NPH}ard{E}val: Dynamic Benchmark on Reasoning Ability of Large Language Models via Complexity Classes",
    author = "Fan, Lizhou  and
      Hua, Wenyue  and
      Li, Lingyao  and
      Ling, Haoyang  and
      Zhang, Yongfeng",
    editor = "Ku, Lun-Wei  and
      Martins, Andre  and
      Srikumar, Vivek",
    booktitle = "Proceedings of the 62nd Annual Meeting of the Association for Computational Linguistics (Volume 1: Long Papers)",
    month = aug,
    year = "2024",
    address = "Bangkok, Thailand",
    publisher = "Association for Computational Linguistics",
    url = "https://aclanthology.org/2024.acl-long.225/",
    doi = "10.18653/v1/2024.acl-long.225",
    pages = "4092--4114"
}

@inproceedings{duchnowski-etal-2025-knapsack,
    title = "A Knapsack by Any Other Name: Presentation impacts {LLM} performance on {NP}-hard problems",
    author = "Duchnowski, Alex  and
      Pavlick, Ellie  and
      Koller, Alexander",
    editor = "Christodoulopoulos, Christos  and
      Chakraborty, Tanmoy  and
      Rose, Carolyn  and
      Peng, Violet",
    booktitle = "Findings of the Association for Computational Linguistics: EMNLP 2025",
    month = nov,
    year = "2025",
    address = "Suzhou, China",
    publisher = "Association for Computational Linguistics",
    url = "https://aclanthology.org/2025.findings-emnlp.352/",
    doi = "10.18653/v1/2025.findings-emnlp.352",
    pages = "6628--6651"
}

@article{heyman2025evaluating,
  title={Evaluating the Systematic Reasoning Abilities of Large Language Models through Graph Coloring},
  author={Heyman, Alex and Zylberberg, Joel},
  journal      = {CoRR},
  volume       = {abs/2502.07087},
  year         = {2025},
}

@inproceedings{hazra2025have,
  title={Have large language models learned to reason? a characterization via 3-sat},
  author={Hazra, Rishi and Venturato, Gabriele and Dos Martires, Pedro Zuidberg and De Raedt, Luc},
  booktitle={Second Conference on Language Modeling},
  year={2025}
}

@inproceedings{
srivastava2026beyondbench,
title={BeyondBench: Contamination-Resistant Evaluation of Reasoning in Language Models},
author={Gaurav Srivastava and Aafiya Shamshad Hussain and Zhenyu Bi and Swastik Roy and Priya Pitre and Meng Lu and Morteza Ziyadi and Xuan Wang},
booktitle={The Fourteenth International Conference on Learning Representations},
year={2026},
url={https://openreview.net/forum?id=mIKqVWGjwI}
}

@misc{meta_llama31_8b,
  author = {{Meta}},
  title = {meta-llama/Llama-3.1-8B},
  year = {2024},
  publisher = {Hugging Face},
  howpublished = {\url{https://huggingface.co/meta-llama/Llama-3.1-8B}},
  note = {Accessed: 2026-07-28}
}

@article{team2026gemma,
  title={Gemma 4 Technical Report},
  author={Team, Gemma and Abd, Sherif El and Aggarwal, Vaibhav and Algayres, Robin and Andreev, Alek and Bachem, Olivier and Ballantyne, Ian and Brick, Cormac and C{\u{a}}rbune, Victor and Casbon, Michelle and others},
    journal      = {CoRR},
  volume       = {abs/2607.02770},
  year         = {2026},
}

@article{agarwal2025gpt,
  title={gpt-oss-120b \& gpt-oss-20b Model Card},
  author={Agarwal, Sandhini and Ahmad, Lama and Ai, Jason and Altman, Sam and Applebaum, Andy and Arbus, Edwin and Arora, Rahul K and Bai, Yu and Baker, Bowen and Bao, Haiming and others},
  journal      = {CoRR},
  volume       = {abs/2508.10925},
  year         = {2025},
}

@article{yang2025qwen3,
  title={Qwen3 Technical Report},
  author={Yang, An and Li, Anfeng and Yang, Baosong and Zhang, Beichen and Hui, Binyuan and Zheng, Bo and Yu, Bowen and Gao, Chang and Huang, Chengen and Lv, Chenxu and others},
  
  journal      = {CoRR},
  volume       = {abs/2505.09388},
  year         = {2025},
}

@article{guo2025deepseek,
  title={Deepseek-R1: Incentivizing Reasoning Capability in LLMs via Reinforcement Learning},
  author={Guo, Daya and Yang, Dejian and Zhang, Haowei and Song, Junxiao and Wang, Peiyi and Zhu, Qihao and Xu, Runxin and Zhang, Ruoyu and Ma, Shirong and Bi, Xiao and others},
  journal      = {CoRR},
  volume       = {abs/2501.12948},
  year         = {2025},
}

@article{DBLP:journals/corr/abs-2605-01120,
  author       = {Jay Bhan and
                  Nicole Nobili and
                  Patrick Langer},
  title        = {New Bounds for Zarankiewicz Numbers via Reinforced {LLM} Evolutionary
                  Search},
  journal      = {CoRR},
  volume       = {abs/2605.01120},
  year         = {2026},
  url          = {https://doi.org/10.48550/arXiv.2605.01120},
  doi          = {10.48550/ARXIV.2605.01120},
  eprinttype   = {arXiv},
  eprint       = {2605.01120},
  bibsource    = {dblp computer science bibliography, https://dblp.org}
}

@misc{openevolve,
  title = {OpenEvolve: an open-source evolutionary coding agent},
  author = {Asankhaya Sharma},
  year = {2025},
  publisher = {GitHub},
  url = {https://github.com/algorithmicsuperintelligence/openevolve}
}

\appendix

\section{Appendix}

\subsection{Hyperparameters}

The ``max\textunderscore tokens'' for each model was set to 32K.
Each model was run using its recommended settings including, hyperparameters, 
system prompt usage,
and other quirks:

\begin{itemize}
    \item \textbf{Llama-3.1-8B-it}
    \begin{itemize}
        \item Temperature: 0.6
        \item Top\textunderscore p: 0.9
        \item System prompt: yes
    \end{itemize}
    \item \textbf{Gemma-4-31B-it}
    \begin{itemize}
        \item Temperature:
        \item Top\textunderscore p: 1
        \item Top\textunderscore k: 0.95
        \item System prompt: yes. As recommended, a ``<|think|>'' tag was added to the system prompt.
    \end{itemize}
    \item \textbf{gpt-oss-120b}
    \begin{itemize}
        \item Temperature: 1
        \item Top\textunderscore p: 1
        \item Top\textunderscore k: 0
        \item System prompt: yes. As recommended, ``reasoning\textunderscore effort = high'' was added to the system prompt.
    \end{itemize}
    \item \textbf{Qwen3-235B-A22B}
    \begin{itemize}
        \item Temperature: 0.7
        \item Top\textunderscore p: 0.8
        \item Top\textunderscore k: 20
        \item Repetition-penalty 1.05
        \item System prompt: yes
    \end{itemize}
    \item \textbf{DeepSeek-R1}
    \begin{itemize}
        \item Temperature: 0.6
        \item Top-p: 0.95
        \item System prompt: no
        \item As recommended, the model was forced to start its response with ``<think>'' to force reasoning.
    \end{itemize}
    
\end{itemize}

\section{Prompts}

System prompts were used to describe the role, output instructions, and necessary definitions. Main prompts were used to provide the task. If a model did not support/recommend system prompts, the system and main prompts would be merged and marked as ``Instructions'' and ``Task,'' respectively. The prompts can be found under ``/ramseygadgetspackage/zero\_shot\_prompts.py'', ``/ramseygadgetspackage/hint\_prompts.py'', and ``/ramseygadgetspackage/multi\_round\_prompts.py'' in the provided code.


\vspace{1em}
\hrule
\vspace{1em}

\end{document}